\pdfoutput=1

\documentclass[11pt]{article}
\usepackage{booktabs}
\usepackage[final]{coling}

\usepackage{times}
\usepackage{latexsym}

\usepackage[T1]{fontenc}

\usepackage[utf8]{inputenc}

\usepackage{microtype}

\usepackage{inconsolata}

\usepackage{graphicx}
\renewcommand{\thefootnote}

\title{From Scarcity to Capability: Empowering Fake News Detection in Low-Resource Languages with LLMs}

\author{Hrithik Majumdar Shibu\textsuperscript{\dag}, Shrestha Datta\textsuperscript{\dag}, {\bf Md. Sumon Miah}, {\bf Nasrullah Sami}, \\
 {\bf Mahruba Sharmin Chowdhury}, {\bf Md. Saiful Islam} \\
Shahjalal University of Science and Technology, Sylhet, Bangladesh \\
\texttt{ \{hrithik11804064, shresthadatta910, iamsumon111, samrevolutionz69\}@gmail.com }\\ 
\texttt{ \{mahruba-cse, saiful-cse\}@sust.edu}   
}

\begin{document}
\maketitle
\begin{abstract}
The rapid spread of fake news presents a significant global challenge, particularly in low-resource languages like Bangla, which lack adequate datasets and detection tools. Although manual fact-checking is accurate, it is expensive and slow to prevent the dissemination of fake news. Addressing this gap, we introduce BanFakeNews-2.0, a robust dataset to enhance Bangla fake news detection. This version includes 11,700 additional, meticulously curated fake news articles validated from credible sources, creating a proportional dataset of 47,000 authentic and 13,000 fake news items across 13 categories. In addition, we created a manually curated independent test set of 460 fake and 540 authentic news items for rigorous evaluation. We invest efforts in collecting fake news from credible sources and manually verified while preserving the linguistic richness. We develop a benchmark system utilizing transformer-based architectures, including fine-tuned Bidirectional Encoder Representations from Transformers variants (F1-87\%) and Large Language Models with Quantized Low-Rank Approximation (F1-89\%), that significantly outperforms traditional methods. 
BanFakeNews-2.0 offers a valuable resource to advance research and application in fake news detection for low-resourced languages.
We publicly release our dataset and model on Github\textsuperscript{1}  
\footnotetext{ \textsuperscript{1} Github: \url{https://github.com/Shibu4064/IndoNLP}} to foster research in this direction.
\end{abstract}

\section{Introduction}

\begin{table}[ht]
    \centering
    \small 
    \begin{tabular}{lcc}
        \hline
        \textbf{Dataset Source} & \textbf{\#FN} & \textbf{\#TN} \\
        \hline
        \citep{bangla_fake} & 4.5K & 10K\\
        \hline
        \citep{ALZAMAN} & 2K & 5k \\
        \hline
        \citep{banfakenews} & 1.3K & 48.6k \\
        \hline
        \citep{gulzar} & 1K & 2.5K \\
        \hline
        \textbf{BanFakeNews-2 (Proposed)}  & 13K & 47k\\
        \hline
    \end{tabular}
    \caption{Overview of existing Bangla fake news datasets. Here \#FN represents No. of fake news and \#TN represents the No. of authentic news dataset}
    \label{tab: data_collection}
\end{table}

\footnotetext{ {\dag} These authors have equal contributions.}

The widespread dissemination of fake news, defined as intentionally misleading information, has become a critical issue in modern society with social consequences. Fake news and misinformation circulate across media channels—from social networks to online news portals—often aiming to mislead and manipulate public opinion. The consequences of such disinformation can range from shaping public opinion on critical matters to catalyzing large-scale societal unrest. For example, during the COVID-19 pandemic, misinformation regarding vaccine safety led to substantial vaccine reluctance \citep{lee2022misinformation, o2020going}. In Bangladesh, the effects of such misinformation have been severe, including incidents of violence and communal discord spurred by false rumors online \citep{ SharifaUmmaShirina_Md.TabiurRahmanProdhan_2020, Bhikkhu}. Moreover, the infodemic—defined as an overabundance of information, including false or misleading details—further complicated efforts to combat COVID-19 globally, as highlighted in studies exploring misinformation trends and mitigation strategies \citep{kouzy2020coronavirus, bridgman2020causes, uddin2021fighting}.  This challenge extends to various content forms, such as articles, images, videos, and memes, amplifying the difficulty of detection \citep{cao2020exploring, Das_2021_ICCV, singh2022predicting, Das_2022_CVPR}. 

Detecting fake news in low-resource languages like Bangla remains challenging due to limited datasets and resources. While English-language fake news detection has progressed, robust datasets for Bangla remain scarce, hindering model development. Although efforts like the BanFakeNews dataset \citep{banfakenews} and others \citep{ALZAMAN} have made initial strides, existing datasets remain limited in size and coverage, and manual fact-checking is impractical at scale. To address these limitations, we present BanFakeNews-2.0, a substantially extended dataset tailored for improved Bangla fake news detection. Building upon BanFakeNews, this new dataset includes 13,000 source-verified fake news articles, forming a balanced collection of 60,000 news items (47,000 authentic, 13,000 fake) across 13 diverse categories compared to the previous largest BanFakeNews dataset. Manually curating an independent test set of 1,000 news articles further enables rigorous model evaluation. Our benchmarks incorporate transformer-based models, such as BERT, and fine-tuned large language models (LLMs) using Quantized Low-Rank Approximation (QLORA). 

BLOOM is a state-of-the-art, open-access large language model that is collaboratively developed by hundreds of researchers and trained on the multilingual ROOTS corpus. It supports 46 natural and 13 programming languages, enabling broad applications and competitive performance across benchmarks \citep{workshop2023bloom176bparameteropenaccessmultilingual}. We observe that our fine-tuned BLOOM 560M model achieves the highest performance, with a macro F1 score of 89. This dataset and benchmark represent a crucial step in advancing fake news detection for low-resource languages like Bangla, providing a foundation for future research and practical applications. Our main contributions include:

\begin{itemize}
    \item We present BanFakeNews-2.0, a significant incremental version of BanFakeNews as shown in Table \ref{tab: data_collection}, while previous research is limited in size and highly imbalanced. We manually collected and validated 60K Bangla news articles, including 13K fake news.
    \item We conducted extensive experiments using traditional linguistic features, transformer-based models like BERT, and LLMs to improve the performance of detecting fake news in Bangla.
    \item We create an independent test set of 1,000 news articles (460 fake, 540 authentic) to ensure rigorous evaluation and cross-comparison of models. 
\end{itemize}

\section{Development of BanFakeNews-2.0}

We focused on data preparation to ensure linguistic richness and dataset diversity with two main objectives: (1) collect verified fake news from diverse sources and domains and (2) enhance dataset variety while minimizing redundancy. Our newly curated dataset comprises approximately 13,000 fake and 47,000 authentic news articles from online news portals and mainstream media. We have collected the misleading or false context type of news mostly from \url{www.jaachai.com} and \url{www.bdfactcheck.com}. These two websites provide a logical and informative explanation of the authenticity of the news published on other sites. So, we have also collected the news mentioned on those two sites from the actual publishing sites and ensured that we avoid duplicates. We have used Python's web-scraping method for automated and accurate collection of category-based news from different online news portals, such as politics, sports, entertainment,  medical, religious, etc. The initial screening has been conducted by evaluating the credibility of sources and verifying claims through fact-checking platforms, authoritative references, or collaborative verification methods. Relevant keywords such as "rumor," "hoax," "viral news," and Bangla-specific terms linked to sensational topics have helped in categorizing the articles. Employing automated web-scraping techniques alongside manual validation ensures data accuracy and quality. Additionally, maintaining a balanced representation of topics, time-frames, and domains has been ensured to create this dataset.

For authentic news, we selected the top 30 Bangladeshi news portals, recognized for their credibility and high readership. For fake news, we gathered content from six major fact-checking platforms that frequently debunk misinformation in Bangladesh, identifying and validating articles as probable fake news for inclusion. To ensure uniqueness, we filtered out duplicates and removed items with over 50\% or 300 words of token overlap, aiming to expand vocabulary diversity and contextual variety. This broad range of content enhances the robustness of our classification system, supporting better generalization across various linguistic styles.

Each article was cross-checked by three annotators to confirm authenticity. Five undergraduate students, guided by detailed source verification protocols, reviewed potentially misleading sources and excluded redundant entries. Note that, we define a verified source of news when the source is at least a person or organization capable of verification of claimed news. When no specific source is available, the reporters or journalists themselves are considered the source of the news. We used majority voting to assign a final label of "fake" or "authentic," achieving a high inter-annotator agreement score of 0.93, indicating strong labelling consistency \citep{fleiss1971measuring}. During dataset analysis, we standardized categories to align with the classifications used in BanFakeNews \citep{banfakenews}, resulting in 13 distinct categories. Categories were assigned based on the classification of the news at its source. If the source did not provide a category, the news was thoroughly read to understand its context and categorized accordingly.
 We focus on increasing the number of fake news articles to reduce the data imbalance, with 500 fake news articles per category. Still, we face challenges in the lifestyle, medical, and religious categories. The final dataset, comprising 60K news articles, is distributed across 13 categories in Table \ref{tab: Table_data}.

\begin{table}[ht]
    \centering
    \small 
    \begin{tabular}{lcc}
        \hline
        \textbf{Category} & \textbf{Authentic} & \textbf{Fake} \\
        \hline
        Politics      & 3141      & 3403      \\

        Miscellaneous & 2218      & 1655      \\
        
        International & 6990      & 1461      \\
        
        Lifestyle     & 901       & 308      \\
        
        Medical       & 112       & 448      \\
        
        Religious     & 118       & 359      \\
        
        Sports        & 6526      & 925      \\
    
        Educational   & 1115      & 808      \\

        Technology    & 843       & 725      \\

        National      & 18708     & 1167      \\
        
        Crime         & 1272      & 720      \\
        
        Entertainment & 2636      & 1441      \\

        Finance       & 1259      & 573      \\
        \hline
    \end{tabular}
    \caption{Statistics of the dataset.}
    \label{tab: Table_data}
\end{table}



\section{Methodologies}

Here, we will outline the methods to create a benchmark model for detecting fake news in Bangla. Our methodologies include traditional linguistic attributes as well as neural networks and transformer-based models.

\subsection{Traditional Approaches} 
We extracted lexical linguistic features using TF-IDF for character n-grams (n = 3,4,5) and word n-grams (n = 1,2,3) similarly as existing works \citep{islam-etal-2022-emonoba}. We applied a Linear Support Vector Machine (SVM) \citep{hearst1998support} to these features for classification. Recognizing the value of semantic information, we experimented with pre-trained word embeddings to represent articles. Specifically, we used Bangla 300-dimensional word vectors pre-trained with FastText on Common Crawl and Wikipedia \citep{banfakenews, romim2022bd}. Finally, we combined all the features with SVM. 

\subsection{Transformer-based BERT Models} 
Encoder-based pre-trained BERT \citep{devlin2018bert} models are exceptional in downstream tasks due to their superior contextual understanding capabilities. We chose five pre-trained model bases: BanglaBERT \citep{bhattacharjee-etal-2022-banglabert} and SagorBERT \citep{Sagor_2020}, which are monolingual, XLM-RoBERTa (XRoBERTa) \citep{DBLP:journals/corr/abs-1911-02116}, multilingual-BERT cased and uncased (m-BERT-c and m-BERT-unc, respectively) \citep{DBLP:journals/corr/abs-1810-04805} which are multilingual. We shuffled the training samples and enforced gradient clipping to fine-tune these models. We utilized the outputs from the last two layers of multi-head attention, subsequently employing a linear layer for classification. We fine-tuned the model using Adam optimizer \citep{kingma2014adam}.

\subsection{Large Language Model} 
Large language models (LLMs) have recently demonstrated impressive linguistic analysis and reasoning abilities. In our experiments, we applied several advanced LLMs to our dataset, including BLOOM 560M \citep{scao2022bloom}, Phi-3 Mini 3.8B \citep{abdin2024phi3}, Stable LM 2 1.6B \citep{bellagente2024stable}, and Llama 3.2 1B \citep{inan2023llamaguardllmbasedinputoutput}.
To fine-tune these models, we employed QLoRA, loading them in 4-bit precision and setting the rank and alpha parameters to 8 and 32, respectively, for trainable adapters. Each model was configured in half-precision floating-point format with normalized 4-bit quantization, using the final token for classification. To manage model complexity and avoid overfitting, alpha is used as a regularization parameter. Its value is adjusted to strike a compromise between underfitting and overfitting \citep{moradi2020survey}. 4-bit quantization \citep{pan2023smoothquantaccurateefficient4bit} is perfect for devices with limited resources or for quicker inference because it drastically reduces model size and increases computing efficiency. Modern quantization methods provide low accuracy loss, allowing for effective deployment with respectable performance. Fine-tuning was optimized through gradient accumulation at each step with a paged Adam 8-bit optimizer\citep{simoulin2024memory}. 


\section{Experimental Setup}
\subsection{Data Preprocessing and Model Validation}
English words and hyperlinks were removed from the dataset. Text normalization, punctuation, and stop-words removal were performed for traditional models.  We have done some pre-processing, including removing NaN values, deleting duplicate rows, etc. As punctuation is essential for capturing context in a sentence, there was no punctuation removal for our LLM experiments.

We validated the models using the holdout method. For this purpose, we split the dataset into train and test sets containing 70\% and 30\%, respectively, following the distribution by the authors of the BanFakeNews \citep{banfakenews} dataset while keeping the same class ratio. We took half of the test split as validation and the rest for testing purposes. This split strikes a practical balance, maintaining sufficient data for each phase while ensuring reliable model evaluation.

\subsection{Baselines} 
In our experimental evaluation, we benchmark our results against two baseline approaches. Firstly, a majority baseline assigns the predominant class label (in this case, 'authentic news') to all articles. The second is a random baseline, which randomly classifies articles as authentic or fake. Table \ref{tab:macro-F1} presents the average precision, recall, and F1-score obtained from 10 random baseline experiments.

\subsection{Experiments} 
For each experiment, we chose the hyperparameters based on the validation set \citep{andonie2019hyperparameter} and evaluated the model on the test set as well as our independent test set. For traditional models, we only trained on the content of the news. For BERTs and LLMs, we trained both on content and headlines while keeping a maximum limit of 512 input tokens. To differentiate the headline and content of each news sample, we added the string `` \ \textbackslash{}\textbackslash{} `` between these.

\section{Result and Analysis}
Table \ref{tab:macro-F1}, describes the performance of various models in terms of Precision (P), Recall (R), and F1 (F1-Score) for both the authentic and fake news classes. Our approach, validated using the independent holdout dataset, yields an unbiased performance measure compared to previous works in Bangla fake news detection. The results indicate high P, R, and F1 scores for the authentic class, with nearly perfect recall. For fake news detection, performance varies by model, reflecting the unique challenges of this classification task.

\begin{table}[ht]
  \centering
  \scriptsize 
  \setlength{\tabcolsep}{6pt}
  \renewcommand{\arraystretch}{1.2} 
  \begin{tabular}{lccccccc}
    \toprule
    \textbf{Model} & \multicolumn{3}{c}{\textbf{Authentic}} & \multicolumn{3}{c}{\textbf{Fake}} & \textbf{Macro} \\
    & \textbf{P} & \textbf{R} & \textbf{F1} & \textbf{P} & \textbf{R} & \textbf{F1} & \textbf{F1} \\
    \midrule
    \multicolumn{8}{l}{\textbf{Baselines}} \\
    Majority & 79 & 100 & 88 & 0 & 0 & 0 & 78 \\
    Random & 79 & 50 & 61 & 21 & 51 & 30 & 63 \\
    \midrule
    \multicolumn{8}{l}{\textbf{Linguistic Features with SVM}} \\
    Unigram(U) & 92 & 95 & 93 & 78 & 70 & 74 & 84 \\
    Bigram(B) & 91 & 95 & 93 & 78 & 67 & 72 & 83 \\
    Trigram(T) & 91 & 88 & 90 & 62 & 69 & 66 & 78 \\
    U+B+T & 92 & 95 & 94 & 79 & 70 & 75 & 85 \\
    C3-Gram(C3) & 96 & 97 & 98 & 80 & 74 & 77 & 86 \\
    C4-Gram(C4) & 97 & 98 & 97 & 79 & 75 & 77 & 86 \\
    C5-Gram(C5) & 96 & 97 & 96 & 81 & 74 & 77 & 86 \\
    C3+C4+C5 & 97 & 98 & 97 & 79 & 75 & 77 & 86 \\
    Embedding & 89 & 98 & 93 & 90 & 57 & 70 & 82 \\
    All Features(All) & 92 & 96 & 94 & 85 & 72 & 78 & 86 \\
    \midrule
    \multicolumn{8}{l}{\textbf{BERT models}} \\
    BanglaBERT & 89 & 99 & 94 & 97 & 53 & 69 & 81 \\
    SagorBERT & 92 & 99 & 95 & 95 & 68 & 79 & 87 \\
    m-BERT-c & 92 & 98 & 95 & 93 & 69 & 79 & 87 \\
    m-BERT-unc & 92 & 98 & 95 & 93 & 70 & 79 & 87 \\ 
    XRoBERTa & 90 & 98 & 94 & 89 & 61 & 72 & 83 \\
    \midrule
    \multicolumn{8}{l}{\textbf{LLMs}} \\
    BLOOM 560M & 92 & 100 & 96 & 99 & 69 & 81 & 89 \\
    Phi 3 mini 3.8B & 90 & 98 & 94 & 92 & 58 & 71 & 83 \\
    Stable LM 2 1.6B & 90 & 98 & 94 & 89 & 61 & 71 & 83 \\
    Llama 3.2 1B & 92 & 99 & 95 & 94 & 66 & 78 & 86
    \\
    \bottomrule
  \end{tabular}
  \caption{Precision (P), Recall (R), and F1 score for each categorical class (Authentic and Fake)}
  \label{tab:macro-F1}
\end{table}

Among word n-grams, unigrams achieved the highest F1 score of 84\%, outperforming bigrams (83\%) and trigrams (78\%). Combining these n-grams resulted in an F1 score of 85\%, demonstrating that multi-gram approach enhances classification accuracy. Character n-grams yielded similar performance; however, combinations of character n-grams did not provide substantial gains. Across experiments, authentic news classification achieved over 90\% in P, R, and F1. However, fake news classification showed greater variability. Traditional SVM models, employing linguistic features, outperformed LLMs and transformers-based models in identifying authentic news. Conversely, LLM-based models excelled in detecting fake news, yielding higher F1 scores. Notably, the transformer models multilingual BERT (m-BERT-unc) and BLOOM achieved an F1 score of 81\% in the fake news class, surpassing the 77\% F1-score achieved by the C3-Gram model. However, traditional models performed slightly better overall, reaching an F1 score of 98\% in the authentic class, compared to the highest F1 score of 96\% for transformers. This discrepancy may stem from the increased volume of fake news in the dataset, posing unique challenges for transformers in handling nuanced distinctions within the fake class. Among the tested transformers, BLOOM and m-BERT-uncased consistently achieved top performance. However, BanglaBERT lagged, exhibiting low P and R for both classes. For linguistic features, character-based models outperformed word-based models in fake news detection. The C3-Gram model surpassed the unigram+bigram+trigram(U+B+T) feature model, showing a 1\%, 4\%, and 2\% higher P, R, and F1, respectively, for fake news. This trend also held for authentic news detection, underscoring the effectiveness of character-level features in handling the nuanced patterns of Bangla fake news.

To assess the generalisability of our models, we evaluated them using a manually curated external test set of 1,000 samples. We tested the top-performing models—the traditional linguistic feature-based SVM and the LLM-based BLOOM—both trained on the BanFakeNews-2.0 dataset, as shown in Table \ref{tab: ablation}. On this external test set, models trained with BanFakeNews-2.0 consistently outperformed those trained on the original BanFakeNews dataset, demonstrating BanFakeNews-2.0’s improved diversity and balance. This enhancement, similar to expanding interview questions to address a wide range of scenarios, equips the models to handle complex and varied data, establishing BanFakeNews-2.0 as a valuable resource for Bangla fake news detection.

\begin{table}
    \centering
       \scriptsize 
         \setlength{\tabcolsep}{6pt}
  \renewcommand{\arraystretch}{1.2} 
     \begin{tabular}{lllc} 
     \toprule
      \textbf{Model} & \textbf{Train dataset} & \textbf{Test dataset} &  \textbf{Mac. F1}  \\ [0.5ex]
     \midrule
     SVM (All) & BanFakeNews & Test (internal) & 74\\
     SVM (All) & BanFakeNews-2.0 & Test (internal) & 86 \\

     SVM (All) & BanFakeNews & Test (external) & 39\\
     SVM (All) & BanFakeNews-2.0 & Test (external) & 91 \\

     BLOOM & BanFakeNews & Test (internal) & 78\\
     BLOOM & BanFakeNews-2.0 & Test (internal) & 89 \\

     BLOOM & BanFakeNews & Test (external) & 29\\
     BLOOM & BanFakeNews-2.0 & Test (external) & 67 \\

     \hline
     
    \end{tabular}
    \caption{Ablation experiments with different train-test combinations of existing BanFakeNews and proposed BanFakeNews-2.0}
    \label{tab: ablation}
\end{table}

\section{Conclusion and Future Works}
The study presents BanFakeNews-2.0, a Bangla fake news dataset with 13K manually annotated articles across 13 categories aimed at improving fake news detection in Bangla. Our evaluation demonstrated that BLOOM and m-BERT-unc models outperformed other models, highlighting the importance of contextually diverse datasets over basic linguistic features for achieving high accuracy. BanFakeNews-2.0 allowed transformer models and LLMs to excel, highlighting the need for diverse datasets and robust detection tools. Future work will focus on enhancing dataset features, refining models, and exploring real-time monitoring. Testing emerging LLMs like Mistral, Minitron, and GPT 4 in zero-shot settings may provide further insights. BanFakeNews-2.0 provides a strong foundation for advancing research in Bangla fake news detection and mitigation.

\section{Limitations}
Generative language models are becoming more human-like, enabling them to imitate authentic news. However, the proposed dataset and pre-trained models may struggle to differentiate advanced fabricated news from upcoming generative models. The low fake news count in some news categories makes it difficult to differentiate. Despite high classification capabilities, the current dataset is imbalanced due to insufficient fake news. A more balanced dataset could improve model capabilities.



\bibliography{acl_latex.bib}

\begin{thebibliography}{34}
\providecommand{\natexlab}[1]{#1}

\bibitem[{Abdin et~al.(2024)Abdin, Jacobs, Awan, Aneja, Awadallah, Awadalla, Bach, Bahree, Bakhtiari, Bao, Behl, Benhaim, Bilenko, Bjorck, Bubeck, Cai, Cai, Mendes, Chen, Chaudhary, Chen, Chen, Chen, Chen, Chopra, Dai, Giorno, de~Rosa, Dixon, Eldan, Fragoso, Iter, Gao, Gao, Gao, Garg, Goswami, Gunasekar, Haider, Hao, Hewett, Huynh, Javaheripi, Jin, Kauffmann, Karampatziakis, Kim, Khademi, Kurilenko, Lee, Lee, Li, Li, Liang, Liden, Liu, Liu, Liu, Lin, Lin, Luo, Madan, Mazzola, Mitra, Modi, Nguyen, Norick, Patra, Perez-Becker, Portet, Pryzant, Qin, Radmilac, Rosset, Roy, Ruwase, Saarikivi, Saied, Salim, Santacroce, Shah, Shang, Sharma, Shukla, Song, Tanaka, Tupini, Wang, Wang, Wang, Wang, Ward, Wang, Witte, Wu, Wyatt, Xiao, Xu, Xu, Xu, Yadav, Yang, Yang, Yang, Yang, Yu, Yuan, Zhang, Zhang, Zhang, Zhang, Zhang, Zhang, Zhang, and Zhou}]{abdin2024phi3}
Marah Abdin, Sam~Ade Jacobs, Ammar~Ahmad Awan, Jyoti Aneja, Ahmed Awadallah, Hany Awadalla, Nguyen Bach, Amit Bahree, Arash Bakhtiari, Jianmin Bao, Harkirat Behl, Alon Benhaim, Misha Bilenko, Johan Bjorck, Sébastien Bubeck, Qin Cai, Martin Cai, Caio César~Teodoro Mendes, Weizhu Chen, Vishrav Chaudhary, Dong Chen, Dongdong Chen, Yen-Chun Chen, Yi-Ling Chen, Parul Chopra, Xiyang Dai, Allie~Del Giorno, Gustavo de~Rosa, Matthew Dixon, Ronen Eldan, Victor Fragoso, Dan Iter, Mei Gao, Min Gao, Jianfeng Gao, Amit Garg, Abhishek Goswami, Suriya Gunasekar, Emman Haider, Junheng Hao, Russell~J. Hewett, Jamie Huynh, Mojan Javaheripi, Xin Jin, Piero Kauffmann, Nikos Karampatziakis, Dongwoo Kim, Mahoud Khademi, Lev Kurilenko, James~R. Lee, Yin~Tat Lee, Yuanzhi Li, Yunsheng Li, Chen Liang, Lars Liden, Ce~Liu, Mengchen Liu, Weishung Liu, Eric Lin, Zeqi Lin, Chong Luo, Piyush Madan, Matt Mazzola, Arindam Mitra, Hardik Modi, Anh Nguyen, Brandon Norick, Barun Patra, Daniel Perez-Becker, Thomas Portet, Reid Pryzant, Heyang
  Qin, Marko Radmilac, Corby Rosset, Sambudha Roy, Olatunji Ruwase, Olli Saarikivi, Amin Saied, Adil Salim, Michael Santacroce, Shital Shah, Ning Shang, Hiteshi Sharma, Swadheen Shukla, Xia Song, Masahiro Tanaka, Andrea Tupini, Xin Wang, Lijuan Wang, Chunyu Wang, Yu~Wang, Rachel Ward, Guanhua Wang, Philipp Witte, Haiping Wu, Michael Wyatt, Bin Xiao, Can Xu, Jiahang Xu, Weijian Xu, Sonali Yadav, Fan Yang, Jianwei Yang, Ziyi Yang, Yifan Yang, Donghan Yu, Lu~Yuan, Chengruidong Zhang, Cyril Zhang, Jianwen Zhang, Li~Lyna Zhang, Yi~Zhang, Yue Zhang, Yunan Zhang, and Xiren Zhou. 2024.
\newblock \href {https://arxiv.org/abs/2404.14219} {Phi-3 technical report: A highly capable language model locally on your phone}.
\newblock \emph{Preprint}, arXiv:2404.14219.

\bibitem[{Al-Zaman and Noman(2023)}]{ALZAMAN}
Md.~Sayeed Al-Zaman and Mridha Md.~Shiblee Noman. 2023.
\newblock \href {https://doi.org/10.1016/j.dib.2023.109439} {A dataset on social media users’ engagement with religious misinformation}.
\newblock \emph{Data in Brief}, 49:109439.

\bibitem[{Andonie(2019)}]{andonie2019hyperparameter}
R{\u{a}}zvan Andonie. 2019.
\newblock Hyperparameter optimization in learning systems.
\newblock \emph{Journal of Membrane Computing}, 1(4):279--291.

\bibitem[{Bellagente et~al.(2024)Bellagente, Tow, Mahan, Phung, Zhuravinskyi, Adithyan, Baicoianu, Brooks, Cooper, Datta et~al.}]{bellagente2024stable}
Marco Bellagente, Jonathan Tow, Dakota Mahan, Duy Phung, Maksym Zhuravinskyi, Reshinth Adithyan, James Baicoianu, Ben Brooks, Nathan Cooper, Ashish Datta, et~al. 2024.
\newblock Stable lm 2 1.6 b technical report.
\newblock \emph{arXiv preprint arXiv:2402.17834}.

\bibitem[{Bhattacharjee et~al.(2022)Bhattacharjee, Hasan, Ahmad, Mubasshir, Islam, Iqbal, Rahman, and Shahriyar}]{bhattacharjee-etal-2022-banglabert}
Abhik Bhattacharjee, Tahmid Hasan, Wasi Ahmad, Kazi~Samin Mubasshir, Md~Saiful Islam, Anindya Iqbal, M.~Sohel Rahman, and Rifat Shahriyar. 2022.
\newblock \href {https://aclanthology.org/2022.findings-naacl.98} {{B}angla{BERT}: Language model pretraining and benchmarks for low-resource language understanding evaluation in {B}angla}.
\newblock In \emph{Findings of the Association for Computational Linguistics: NAACL 2022}, pages 1318--1327, Seattle, United States. Association for Computational Linguistics.

\bibitem[{Bhikkhu(2014)}]{Bhikkhu}
Pragyananda Bhikkhu. 2014.
\newblock \href {https://en.prothomalo.com/opinion/Who-will-be-tried-for-Ramu-destruction} {Who will be tried for ramu destruction?}
\newblock Published: 30 Sep 2014, 16: 58.

\bibitem[{Bridgman et~al.(2020)Bridgman, Merkley, Loewen, Owen, Ruths, Teichmann, and Zhilin}]{bridgman2020causes}
Aengus Bridgman, Eric Merkley, Peter~John Loewen, Taylor Owen, Derek Ruths, Lisa Teichmann, and Oleg Zhilin. 2020.
\newblock The causes and consequences of covid-19 misperceptions: Understanding the role of news and social media.
\newblock \emph{Harvard Kennedy School Misinformation Review}, 1(3).

\bibitem[{Cao et~al.(2020)Cao, Qi, Sheng, Yang, Guo, and Li}]{cao2020exploring}
Juan Cao, Peng Qi, Qiang Sheng, Tianyun Yang, Junbo Guo, and Jintao Li. 2020.
\newblock Exploring the role of visual content in fake news detection.
\newblock \emph{Disinformation, Misinformation, and Fake News in Social Media: Emerging Research Challenges and Opportunities}, pages 141--161.

\bibitem[{Conneau et~al.(2019)Conneau, Khandelwal, Goyal, Chaudhary, Wenzek, Guzm{\'{a}}n, Grave, Ott, Zettlemoyer, and Stoyanov}]{DBLP:journals/corr/abs-1911-02116}
Alexis Conneau, Kartikay Khandelwal, Naman Goyal, Vishrav Chaudhary, Guillaume Wenzek, Francisco Guzm{\'{a}}n, Edouard Grave, Myle Ott, Luke Zettlemoyer, and Veselin Stoyanov. 2019.
\newblock \href {https://arxiv.org/abs/1911.02116} {Unsupervised cross-lingual representation learning at scale}.
\newblock \emph{CoRR}, abs/1911.02116.

\bibitem[{Das et~al.(2022)Das, Islam, and Amin}]{Das_2022_CVPR}
Sowmen Das, Md.~Saiful Islam, and Md.~Ruhul Amin. 2022.
\newblock Gca-net: Utilizing gated context attention for improving image forgery localization and detection.
\newblock In \emph{Proceedings of the IEEE/CVF Conference on Computer Vision and Pattern Recognition (CVPR) Workshops}, pages 81--90.

\bibitem[{Das et~al.(2021)Das, Seferbekov, Datta, Islam, and Amin}]{Das_2021_ICCV}
Sowmen Das, Selim Seferbekov, Arup Datta, Md.~Saiful Islam, and Md.~Ruhul Amin. 2021.
\newblock Towards solving the deepfake problem: An analysis on improving deepfake detection using dynamic face augmentation.
\newblock In \emph{Proceedings of the IEEE/CVF International Conference on Computer Vision (ICCV) Workshops}, pages 3776--3785.

\bibitem[{Devlin et~al.(2018{\natexlab{a}})Devlin, Chang, Lee, and Toutanova}]{devlin2018bert}
Jacob Devlin, Ming-Wei Chang, Kenton Lee, and Kristina Toutanova. 2018{\natexlab{a}}.
\newblock Bert: Pre-training of deep bidirectional transformers for language understanding.
\newblock \emph{arXiv preprint arXiv:1810.04805}.

\bibitem[{Devlin et~al.(2018{\natexlab{b}})Devlin, Chang, Lee, and Toutanova}]{DBLP:journals/corr/abs-1810-04805}
Jacob Devlin, Ming{-}Wei Chang, Kenton Lee, and Kristina Toutanova. 2018{\natexlab{b}}.
\newblock \href {https://arxiv.org/abs/1810.04805} {{BERT:} pre-training of deep bidirectional transformers for language understanding}.
\newblock \emph{CoRR}, abs/1810.04805.

\bibitem[{Fleiss(1971)}]{fleiss1971measuring}
Joseph~L Fleiss. 1971.
\newblock Measuring nominal scale agreement among many raters.
\newblock \emph{Psychological bulletin}, 76(5):378.

\bibitem[{Hearst et~al.(1998)Hearst, Dumais, Osuna, Platt, and Scholkopf}]{hearst1998support}
Marti~A. Hearst, Susan~T Dumais, Edgar Osuna, John Platt, and Bernhard Scholkopf. 1998.
\newblock Support vector machines.
\newblock \emph{IEEE Intelligent Systems and their applications}, 13(4):18--28.

\bibitem[{Hossain et~al.(2020)Hossain, Rahman, Islam, and Kar}]{banfakenews}
Md~Zobaer Hossain, Md~Ashraful Rahman, Md~Saiful Islam, and Sudipta Kar. 2020.
\newblock \href {https://aclanthology.org/2020.lrec-1.349} {{B}an{F}ake{N}ews: A dataset for detecting fake news in {B}angla}.
\newblock In \emph{Proceedings of the Twelfth Language Resources and Evaluation Conference}, pages 2862--2871, Marseille, France. European Language Resources Association.

\bibitem[{Hussain et~al.(2020)Hussain, Rashidul~Hasan, Rahman, Protim, and Al~Hasan}]{gulzar}
Md~Gulzar Hussain, Md~Rashidul~Hasan, Mahmuda Rahman, Joy Protim, and Sakib Al~Hasan. 2020.
\newblock \href {https://doi.org/10.1109/iCCECE49321.2020.9231167} {Detection of bangla fake news using mnb and svm classifier}.
\newblock In \emph{2020 International Conference on Computing, Electronics \& Communications Engineering (iCCECE)}, pages 81--85.

\bibitem[{Inan et~al.(2023)Inan, Upasani, Chi, Rungta, Iyer, Mao, Tontchev, Hu, Fuller, Testuggine, and Khabsa}]{inan2023llamaguardllmbasedinputoutput}
Hakan Inan, Kartikeya Upasani, Jianfeng Chi, Rashi Rungta, Krithika Iyer, Yuning Mao, Michael Tontchev, Qing Hu, Brian Fuller, Davide Testuggine, and Madian Khabsa. 2023.
\newblock \href {https://arxiv.org/abs/2312.06674} {Llama guard: Llm-based input-output safeguard for human-ai conversations}.
\newblock \emph{Preprint}, arXiv:2312.06674.

\bibitem[{Islam et~al.(2022)Islam, Yuvraz, Islam, and Hassan}]{islam-etal-2022-emonoba}
Khondoker~Ittehadul Islam, Tanvir Yuvraz, Md~Saiful Islam, and Enamul Hassan. 2022.
\newblock \href {https://aclanthology.org/2022.aacl-short.17} {{E}mo{N}o{B}a: A dataset for analyzing fine-grained emotions on noisy {B}angla texts}.
\newblock In \emph{Proceedings of the 2nd Conference of the Asia-Pacific Chapter of the Association for Computational Linguistics and the 12th International Joint Conference on Natural Language Processing (Volume 2: Short Papers)}, pages 128--134, Online only. Association for Computational Linguistics.

\bibitem[{Kingma and Ba(2014)}]{kingma2014adam}
Diederik~P Kingma and Jimmy Ba. 2014.
\newblock Adam: A method for stochastic optimization.
\newblock \emph{arXiv preprint arXiv:1412.6980}.

\bibitem[{Kouzy et~al.(2020)Kouzy, Abi~Jaoude, Kraitem, El~Alam, Karam, Adib, Zarka, Traboulsi, Akl, and Baddour}]{kouzy2020coronavirus}
Ramez Kouzy, Joseph Abi~Jaoude, Afif Kraitem, Molly~B El~Alam, Basil Karam, Elio Adib, Jabra Zarka, Cindy Traboulsi, Elie~W Akl, and Khalil Baddour. 2020.
\newblock Coronavirus goes viral: quantifying the covid-19 misinformation epidemic on twitter.
\newblock \emph{Cureus}, 12(3).

\bibitem[{Lee et~al.(2022)Lee, Sun, Jang, and Connelly}]{lee2022misinformation}
Sun~Kyong Lee, Juhyung Sun, Seulki Jang, and Shane Connelly. 2022.
\newblock Misinformation of covid-19 vaccines and vaccine hesitancy.

\bibitem[{Moradi et~al.(2020)Moradi, Berangi, and Minaei}]{moradi2020survey}
Reza Moradi, Reza Berangi, and Behrouz Minaei. 2020.
\newblock A survey of regularization strategies for deep models.
\newblock \emph{Artificial Intelligence Review}, 53(6):3947--3986.

\bibitem[{O’Connor and Murphy(2020)}]{o2020going}
Cathal O’Connor and Michelle Murphy. 2020.
\newblock Going viral: doctors must tackle fake news in the covid-19 pandemic.
\newblock \emph{Bmj}, 369(10.1136).

\bibitem[{Pan et~al.(2023)Pan, Wang, Zheng, Li, Wang, and Feng}]{pan2023smoothquantaccurateefficient4bit}
Jiayi Pan, Chengcan Wang, Kaifu Zheng, Yangguang Li, Zhenyu Wang, and Bin Feng. 2023.
\newblock \href {https://arxiv.org/abs/2312.03788} {Smoothquant+: Accurate and efficient 4-bit post-training weightquantization for llm}.
\newblock \emph{Preprint}, arXiv:2312.03788.

\bibitem[{Romim et~al.(2022)Romim, Ahmed, Islam, Sharma, Talukder, and Amin}]{romim2022bd}
Nauros Romim, Mosahed Ahmed, Md~Saiful Islam, Arnab~Sen Sharma, Hriteshwar Talukder, and Mohammad~Ruhul Amin. 2022.
\newblock Bd-shs: A benchmark dataset for learning to detect online bangla hate speech in different social contexts.
\newblock In \emph{Proceedings of the Thirteenth Language Resources and Evaluation Conference}, pages 5153--5162.

\bibitem[{SadikAlJarif(2022)}]{bangla_fake}
SadikAlJarif. 2022.
\newblock bangla fake news dataset.
\newblock \url{https://www.kaggle.com/datasets/sadikaljarif/bangla-fake-news-detection-dataset?select=final_bn_data.csv}.

\bibitem[{Sarker(2020)}]{Sagor_2020}
Sagor Sarker. 2020.
\newblock \href {https://github.com/sagorbrur/bangla-bert} {Banglabert: Bengali mask language model for bengali language understanding}.

\bibitem[{Scao et~al.(2022)Scao, Fan, Akiki, Pavlick, Ili{\'c}, Hesslow, Castagn{\'e}, Luccioni, Yvon, Gall{\'e} et~al.}]{scao2022bloom}
Teven~Le Scao, Angela Fan, Christopher Akiki, Ellie Pavlick, Suzana Ili{\'c}, Daniel Hesslow, Roman Castagn{\'e}, Alexandra~Sasha Luccioni, Fran{\c{c}}ois Yvon, Matthias Gall{\'e}, et~al. 2022.
\newblock Bloom: A 176b-parameter open-access multilingual language model.
\newblock \emph{arXiv preprint arXiv:2211.05100}.

\bibitem[{Shirina and Prodhan(2020)}]{SharifaUmmaShirina_Md.TabiurRahmanProdhan_2020}
Sharifa~Umma Shirina and Md. Tabiur~Rahman Prodhan. 2020.
\newblock \href {https://doi.org/10.34257/GJHSSAVOL20IS17PG11} {Spreading fake news in the virtual realm in bangladesh: Assessment of impact}.
\newblock \emph{Global Journal of Human-Social Science}, 20(A17):11–25.

\bibitem[{Simoulin et~al.(2024)Simoulin, Park, Liu, and Yang}]{simoulin2024memory}
Antoine Simoulin, Namyong Park, Xiaoyi Liu, and Grey Yang. 2024.
\newblock Memory-efficient fine-tuning of transformers via token selection.
\newblock In \emph{Proceedings of the 2024 Conference on Empirical Methods in Natural Language Processing}, pages 21565--21580.

\bibitem[{Singh and Sharma(2022)}]{singh2022predicting}
Bhuvanesh Singh and Dilip~Kumar Sharma. 2022.
\newblock Predicting image credibility in fake news over social media using multi-modal approach.
\newblock \emph{Neural Computing and Applications}, 34(24):21503--21517.

\bibitem[{Uddin et~al.(2021)Uddin, Reza, Islam, Ahsan, and Amin}]{uddin2021fighting}
Borhan Uddin, Nahid Reza, Md~Saiful Islam, Hasib Ahsan, and Mohammad~Ruhul Amin. 2021.
\newblock Fighting against fake news during pandemic era: Does providing related news help student internet users to detect covid-19 misinformation?
\newblock In \emph{Proceedings of the 13th ACM Web Science Conference 2021}, pages 178--186.

\bibitem[{Workshop et~al.(2023)Workshop, :, Scao, Fan, Akiki, Pavlick, Ilić, Hesslow, Castagné, Luccioni, Yvon, Gallé, Tow, Rush, Biderman, Webson, Ammanamanchi, Wang, Sagot, Muennighoff, del Moral, Ruwase, Bawden, Bekman, McMillan-Major, Beltagy, Nguyen, Saulnier, Tan, Suarez, Sanh, Laurençon, Jernite, Launay, Mitchell, Raffel, Gokaslan, Simhi, Soroa, Aji, Alfassy, Rogers, Nitzav, Xu, Mou, Emezue, Klamm, Leong, van Strien, Adelani, Radev, Ponferrada, Levkovizh, Kim, Natan, Toni, Dupont, Kruszewski, Pistilli, Elsahar, Benyamina, Tran, Yu, Abdulmumin, Johnson, Gonzalez-Dios, de~la Rosa, Chim, Dodge, Zhu, Chang, Frohberg, Tobing, Bhattacharjee, Almubarak, Chen, Lo, Werra, Weber, Phan, allal, Tanguy, Dey, Muñoz, Masoud, Grandury, Šaško, Huang, Coavoux, Singh, Jiang, Vu, Jauhar, Ghaleb, Subramani, Kassner, Khamis, Nguyen, Espejel, de~Gibert, Villegas, Henderson, Colombo, Amuok, Lhoest, Harliman, Bommasani, López, Ribeiro, Osei, Pyysalo, Nagel, Bose, Muhammad, Sharma, Longpre, Nikpoor, Silberberg, Pai,
  Zink, Torrent, Schick, Thrush, Danchev, Nikoulina, Laippala, Lepercq, Prabhu, Alyafeai, Talat, Raja, Heinzerling, Si, Taşar, Salesky, Mielke, Lee, Sharma, Santilli, Chaffin, Stiegler, Datta, Szczechla, Chhablani, Wang, Pandey, Strobelt, Fries, Rozen, Gao, Sutawika, Bari, Al-shaibani, Manica, Nayak, Teehan, Albanie, Shen, Ben-David, Bach, Kim, Bers, Fevry, Neeraj, Thakker, Raunak, Tang, Yong, Sun, Brody, Uri, Tojarieh, Roberts, Chung, Tae, Phang, Press, Li, Narayanan, Bourfoune, Casper, Rasley, Ryabinin, Mishra, Zhang, Shoeybi, Peyrounette, Patry, Tazi, Sanseviero, von Platen, Cornette, Lavallée, Lacroix, Rajbhandari, Gandhi, Smith, Requena, Patil, Dettmers, Baruwa, Singh, Cheveleva, Ligozat, Subramonian, Névéol, Lovering, Garrette, Tunuguntla, Reiter, Taktasheva, Voloshina, Bogdanov, Winata, Schoelkopf, Kalo, Novikova, Forde, Clive, Kasai, Kawamura, Hazan, Carpuat, Clinciu, Kim, Cheng, Serikov, Antverg, van~der Wal, Zhang, Zhang, Gehrmann, Mirkin, Pais, Shavrina, Scialom, Yun, Limisiewicz, Rieser,
  Protasov, Mikhailov, Pruksachatkun, Belinkov, Bamberger, Kasner, Rueda, Pestana, Feizpour, Khan, Faranak, Santos, Hevia, Unldreaj, Aghagol, Abdollahi, Tammour, HajiHosseini, Behroozi, Ajibade, Saxena, Ferrandis, McDuff, Contractor, Lansky, David, Kiela, Nguyen, Tan, Baylor, Ozoani, Mirza, Ononiwu, Rezanejad, Jones, Bhattacharya, Solaiman, Sedenko, Nejadgholi, Passmore, Seltzer, Sanz, Dutra, Samagaio, Elbadri, Mieskes, Gerchick, Akinlolu, McKenna, Qiu, Ghauri, Burynok, Abrar, Rajani, Elkott, Fahmy, Samuel, An, Kromann, Hao, Alizadeh, Shubber, Wang, Roy, Viguier, Le, Oyebade, Le, Yang, Nguyen, Kashyap, Palasciano, Callahan, Shukla, Miranda-Escalada, Singh, Beilharz, Wang, Brito, Zhou, Jain, Xu, Fourrier, Periñán, Molano, Yu, Manjavacas, Barth, Fuhrimann, Altay, Bayrak, Burns, Vrabec, Bello, Dash, Kang, Giorgi, Golde, Posada, Sivaraman, Bulchandani, Liu, Shinzato, de~Bykhovetz, Takeuchi, Pàmies, Castillo, Nezhurina, Sänger, Samwald, Cullan, Weinberg, Wolf, Mihaljcic, Liu, Freidank, Kang, Seelam, Dahlberg,
  Broad, Muellner, Fung, Haller, Chandrasekhar, Eisenberg, Martin, Canalli, Su, Su, Cahyawijaya, Garda, Deshmukh, Mishra, Kiblawi, Ott, Sang-aroonsiri, Kumar, Schweter, Bharati, Laud, Gigant, Kainuma, Kusa, Labrak, Bajaj, Venkatraman, Xu, Xu, Xu, Tan, Xie, Ye, Bras, Belkada, and Wolf}]{workshop2023bloom176bparameteropenaccessmultilingual}
BigScience Workshop, :, Teven~Le Scao, Angela Fan, Christopher Akiki, Ellie Pavlick, Suzana Ilić, Daniel Hesslow, Roman Castagné, Alexandra~Sasha Luccioni, François Yvon, Matthias Gallé, Jonathan Tow, Alexander~M. Rush, Stella Biderman, Albert Webson, Pawan~Sasanka Ammanamanchi, Thomas Wang, Benoît Sagot, Niklas Muennighoff, Albert~Villanova del Moral, Olatunji Ruwase, Rachel Bawden, Stas Bekman, Angelina McMillan-Major, Iz~Beltagy, Huu Nguyen, Lucile Saulnier, Samson Tan, Pedro~Ortiz Suarez, Victor Sanh, Hugo Laurençon, Yacine Jernite, Julien Launay, Margaret Mitchell, Colin Raffel, Aaron Gokaslan, Adi Simhi, Aitor Soroa, Alham~Fikri Aji, Amit Alfassy, Anna Rogers, Ariel~Kreisberg Nitzav, Canwen Xu, Chenghao Mou, Chris Emezue, Christopher Klamm, Colin Leong, Daniel van Strien, David~Ifeoluwa Adelani, Dragomir Radev, Eduardo~González Ponferrada, Efrat Levkovizh, Ethan Kim, Eyal~Bar Natan, Francesco~De Toni, Gérard Dupont, Germán Kruszewski, Giada Pistilli, Hady Elsahar, Hamza Benyamina, Hieu Tran,
  Ian Yu, Idris Abdulmumin, Isaac Johnson, Itziar Gonzalez-Dios, Javier de~la Rosa, Jenny Chim, Jesse Dodge, Jian Zhu, Jonathan Chang, Jörg Frohberg, Joseph Tobing, Joydeep Bhattacharjee, Khalid Almubarak, Kimbo Chen, Kyle Lo, Leandro~Von Werra, Leon Weber, Long Phan, Loubna~Ben allal, Ludovic Tanguy, Manan Dey, Manuel~Romero Muñoz, Maraim Masoud, María Grandury, Mario Šaško, Max Huang, Maximin Coavoux, Mayank Singh, Mike Tian-Jian Jiang, Minh~Chien Vu, Mohammad~A. Jauhar, Mustafa Ghaleb, Nishant Subramani, Nora Kassner, Nurulaqilla Khamis, Olivier Nguyen, Omar Espejel, Ona de~Gibert, Paulo Villegas, Peter Henderson, Pierre Colombo, Priscilla Amuok, Quentin Lhoest, Rheza Harliman, Rishi Bommasani, Roberto~Luis López, Rui Ribeiro, Salomey Osei, Sampo Pyysalo, Sebastian Nagel, Shamik Bose, Shamsuddeen~Hassan Muhammad, Shanya Sharma, Shayne Longpre, Somaieh Nikpoor, Stanislav Silberberg, Suhas Pai, Sydney Zink, Tiago~Timponi Torrent, Timo Schick, Tristan Thrush, Valentin Danchev, Vassilina Nikoulina,
  Veronika Laippala, Violette Lepercq, Vrinda Prabhu, Zaid Alyafeai, Zeerak Talat, Arun Raja, Benjamin Heinzerling, Chenglei Si, Davut~Emre Taşar, Elizabeth Salesky, Sabrina~J. Mielke, Wilson~Y. Lee, Abheesht Sharma, Andrea Santilli, Antoine Chaffin, Arnaud Stiegler, Debajyoti Datta, Eliza Szczechla, Gunjan Chhablani, Han Wang, Harshit Pandey, Hendrik Strobelt, Jason~Alan Fries, Jos Rozen, Leo Gao, Lintang Sutawika, M~Saiful Bari, Maged~S. Al-shaibani, Matteo Manica, Nihal Nayak, Ryan Teehan, Samuel Albanie, Sheng Shen, Srulik Ben-David, Stephen~H. Bach, Taewoon Kim, Tali Bers, Thibault Fevry, Trishala Neeraj, Urmish Thakker, Vikas Raunak, Xiangru Tang, Zheng-Xin Yong, Zhiqing Sun, Shaked Brody, Yallow Uri, Hadar Tojarieh, Adam Roberts, Hyung~Won Chung, Jaesung Tae, Jason Phang, Ofir Press, Conglong Li, Deepak Narayanan, Hatim Bourfoune, Jared Casper, Jeff Rasley, Max Ryabinin, Mayank Mishra, Minjia Zhang, Mohammad Shoeybi, Myriam Peyrounette, Nicolas Patry, Nouamane Tazi, Omar Sanseviero, Patrick von
  Platen, Pierre Cornette, Pierre~François Lavallée, Rémi Lacroix, Samyam Rajbhandari, Sanchit Gandhi, Shaden Smith, Stéphane Requena, Suraj Patil, Tim Dettmers, Ahmed Baruwa, Amanpreet Singh, Anastasia Cheveleva, Anne-Laure Ligozat, Arjun Subramonian, Aurélie Névéol, Charles Lovering, Dan Garrette, Deepak Tunuguntla, Ehud Reiter, Ekaterina Taktasheva, Ekaterina Voloshina, Eli Bogdanov, Genta~Indra Winata, Hailey Schoelkopf, Jan-Christoph Kalo, Jekaterina Novikova, Jessica~Zosa Forde, Jordan Clive, Jungo Kasai, Ken Kawamura, Liam Hazan, Marine Carpuat, Miruna Clinciu, Najoung Kim, Newton Cheng, Oleg Serikov, Omer Antverg, Oskar van~der Wal, Rui Zhang, Ruochen Zhang, Sebastian Gehrmann, Shachar Mirkin, Shani Pais, Tatiana Shavrina, Thomas Scialom, Tian Yun, Tomasz Limisiewicz, Verena Rieser, Vitaly Protasov, Vladislav Mikhailov, Yada Pruksachatkun, Yonatan Belinkov, Zachary Bamberger, Zdeněk Kasner, Alice Rueda, Amanda Pestana, Amir Feizpour, Ammar Khan, Amy Faranak, Ana Santos, Anthony Hevia, Antigona
  Unldreaj, Arash Aghagol, Arezoo Abdollahi, Aycha Tammour, Azadeh HajiHosseini, Bahareh Behroozi, Benjamin Ajibade, Bharat Saxena, Carlos~Muñoz Ferrandis, Daniel McDuff, Danish Contractor, David Lansky, Davis David, Douwe Kiela, Duong~A. Nguyen, Edward Tan, Emi Baylor, Ezinwanne Ozoani, Fatima Mirza, Frankline Ononiwu, Habib Rezanejad, Hessie Jones, Indrani Bhattacharya, Irene Solaiman, Irina Sedenko, Isar Nejadgholi, Jesse Passmore, Josh Seltzer, Julio~Bonis Sanz, Livia Dutra, Mairon Samagaio, Maraim Elbadri, Margot Mieskes, Marissa Gerchick, Martha Akinlolu, Michael McKenna, Mike Qiu, Muhammed Ghauri, Mykola Burynok, Nafis Abrar, Nazneen Rajani, Nour Elkott, Nour Fahmy, Olanrewaju Samuel, Ran An, Rasmus Kromann, Ryan Hao, Samira Alizadeh, Sarmad Shubber, Silas Wang, Sourav Roy, Sylvain Viguier, Thanh Le, Tobi Oyebade, Trieu Le, Yoyo Yang, Zach Nguyen, Abhinav~Ramesh Kashyap, Alfredo Palasciano, Alison Callahan, Anima Shukla, Antonio Miranda-Escalada, Ayush Singh, Benjamin Beilharz, Bo~Wang, Caio Brito,
  Chenxi Zhou, Chirag Jain, Chuxin Xu, Clémentine Fourrier, Daniel~León Periñán, Daniel Molano, Dian Yu, Enrique Manjavacas, Fabio Barth, Florian Fuhrimann, Gabriel Altay, Giyaseddin Bayrak, Gully Burns, Helena~U. Vrabec, Imane Bello, Ishani Dash, Jihyun Kang, John Giorgi, Jonas Golde, Jose~David Posada, Karthik~Rangasai Sivaraman, Lokesh Bulchandani, Lu~Liu, Luisa Shinzato, Madeleine~Hahn de~Bykhovetz, Maiko Takeuchi, Marc Pàmies, Maria~A Castillo, Marianna Nezhurina, Mario Sänger, Matthias Samwald, Michael Cullan, Michael Weinberg, Michiel~De Wolf, Mina Mihaljcic, Minna Liu, Moritz Freidank, Myungsun Kang, Natasha Seelam, Nathan Dahlberg, Nicholas~Michio Broad, Nikolaus Muellner, Pascale Fung, Patrick Haller, Ramya Chandrasekhar, Renata Eisenberg, Robert Martin, Rodrigo Canalli, Rosaline Su, Ruisi Su, Samuel Cahyawijaya, Samuele Garda, Shlok~S Deshmukh, Shubhanshu Mishra, Sid Kiblawi, Simon Ott, Sinee Sang-aroonsiri, Srishti Kumar, Stefan Schweter, Sushil Bharati, Tanmay Laud, Théo Gigant, Tomoya
  Kainuma, Wojciech Kusa, Yanis Labrak, Yash~Shailesh Bajaj, Yash Venkatraman, Yifan Xu, Yingxin Xu, Yu~Xu, Zhe Tan, Zhongli Xie, Zifan Ye, Mathilde Bras, Younes Belkada, and Thomas Wolf. 2023.
\newblock \href {https://arxiv.org/abs/2211.05100} {Bloom: A 176b-parameter open-access multilingual language model}.
\newblock \emph{Preprint}, arXiv:2211.05100.

\end{thebibliography}

\appendix

\section{Appendix}
\label{sec:appendix}
\centering
\par{Authentic News Sources}
\begin{table}[ht]
    \centering
    \small 
    \begin{tabular}{lcc}
        \hline
        \textbf{Domain} & \textbf{Count} \\
        \hline
         \url{www.kalerkantho.com}      & 4491      \\

         \url{www.jagonews24.com}    & 4426   \\
        
         \url{www.banglanews24.com} & 4035  \\
        
        \url{www.banglatribune.com}     &  3696     \\
        
         \url{www.jugantor.com}     & 2835     \\
        
         \url{www.dhakatimes24.com}     & 2654      \\
        
          \url{www.ittefaq.com.bd}      &  2589     \\
               
         \url{www.somoynews.tv}   &  2552     \\

         \url{www.dailynayadiganta.com}    & 2371      \\

         \url{www.bangla.bdnews24.com}      &  2365   \\
        
         \url{www.prothomalo.com}        & 2350     \\
        
         \url{www.bd24live.com} & 2335      \\

           \url{www.risingbd.com}       &  2220   \\

          \url{www.dailyjanakantha.com}   & 1531 \\

          \url{www.bd-pratidin.com}   &  1421 \\

          \url{www.channelionline.com} & 1401 \\

          \url{www.samakal.com} & 1372 \\

          \url{www.independent24.com} & 1220 \\

          \url{www.rtnn.net} & 1149 \\

          \url{www.bangla.thereport24.com} & 859 \\

          \url{www.mzamin.com} & 785 \\

          \url{www.bhorerkagoj.net} & 21 \\
           
        \hline
    \end{tabular}
    \label{tab: Table_data_1}
    \caption{Detailed statistics of the collected authentic news with the domain URL}
\end{table}

\centering
\par{Fake News Sources}
\begin{table}[ht]
    \centering
    \small 
    \begin{tabular}{lcc}
        \hline
        \textbf{Domain} & \textbf{Count} \\
        \hline
         \url{www.boombd.com/fake-news}      & 321      \\

         \url{www.anandabazar.com/topic/fake-news}    & 192     \\
        
         \url{www.jachai.org/fact-checks} & 345  \\
        
        \url{www.bangla.hindustantimes.com/fake}     & 272     \\
        
         \url{www.earki.co}     & 231     \\
        
         \url{www.balerkontho.net/2020/03/}     & 138      \\
        
          \url{www.prothom1alu.blogspot.com}      & 154     \\
               
         \url{www.motikontho.wordpress.com}   & 271     \\

         \url{www.bengalbeats.com}    & 204      \\

         \url{www.shadhinbangla24.com.bd}      & 291   \\
        
         \url{www.bengaliviralnews.com}        & 268     \\
        
         \url{www.shawdeshbhumi.com} & 373      \\

           \url{www.bdexclusivenews.blogspot.com}       & 312   \\

          \url{www.banglainsider.com}   & 277 \\

          \url{www.bd-pratidin.com}   &  293 \\

          \url{www.dailyinqilab.com} & 191 \\

          \url{www.bangla.dhakatribune.com} & 267 \\

        \hline
    \end{tabular}
    \label{tab: Table_data_2}
    \caption{Detailed statistics of the collected fake news with the domain URL}
\end{table}

\end{document}